\documentclass[preprint,12pt]{elsarticle}

\usepackage{amssymb}
\usepackage{amsmath}
\usepackage[normalem]{ulem}
\usepackage{lscape}
\usepackage{longtable}
\usepackage[table,xcdraw]{xcolor}
\usepackage{graphicx}
\usepackage{tabularx}
\usepackage{adjustbox}
\usepackage{wrapfig}
\usepackage{rotating}
\usepackage{epstopdf}
\usepackage{multirow}
\usepackage{placeins}
\usepackage{silence}
\usepackage{geometry}
\usepackage{array}
\usepackage{subcaption}
\usepackage{xurl}

\journal{Applied Energy}

\begin{document}

\begin{frontmatter}

\title{Leveraging Asynchronous Cross-border Market Data for Improved Day-Ahead Electricity Price Forecasting in European Markets}

\author[KUL]{Maria Margarida Mascarenhas\footnote{Both Maria Margarida Mascarenhas and Jilles De Blauwe contributed equally to the paper.}}
\ead{margarida.mascarenhas@kuleuven.be}

\author[DTU]{Jilles De Blauwe\corref{cor2}}%
\ead{jcdbl@dtu.dk}

\author[KTH]{Mikael Amelin}%
\ead{amelin@kth.se}

\author[KUL]{Hussain Kazmi}
\ead{hussain.kazmi@kuleuven.be}

\affiliation[KUL]{organization={KU Leuven, Belgium}}
\affiliation[DTU]{organization={Technical University of Denmark, Denmark}}
\affiliation[KTH]{organization={KTH Royal Institute of Technology, Sweden}}

\begin{abstract}

Accurate short-term electricity price forecasting is crucial for strategically scheduling demand and generation bids in day-ahead markets. While data-driven techniques have shown considerable prowess in achieving high forecast accuracy in recent years, they rely heavily on the quality of input covariates. In this paper, we investigate whether asynchronously published prices as a result of differing gate closure times (GCTs) in some bidding zones can improve forecasting accuracy in other markets with later GCTs. Using a state-of-the-art ensemble of models, we show significant improvements of 22\% and 9\% in forecast accuracy in the Belgian (BE) and Swedish bidding zones (SE3) respectively, when including price data from interconnected markets with earlier GCT (Germany-Luxembourg, Austria, and Switzerland). This improvement holds for both general as well as extreme market conditions. Our analysis also yields further important insights: frequent model recalibration is necessary for maximum accuracy but comes at substantial additional computational costs, and using data from more markets does not always lead to better performance - a fact we delve deeper into with interpretability analysis of the forecast models. Overall, these findings provide valuable guidance for market participants and decision-makers aiming to optimize bidding strategies within increasingly interconnected and volatile European energy markets.

\end{abstract}


\end{frontmatter}

\section{Introduction}\label{Sec1:Introduction}

\subsection{The Electricity Spot Market}\label{subsec1.1:Intro_Spot_market}
In most European countries, electricity has been a freely traded commodity ever since the liberalization of the sector starting in 1996. The day-ahead market (DAM), also referred to as the spot market, is among the most actively traded markets in many countries, where electricity to be delivered the next day is traded. Usually, one spot market is organized per bidding zone, facilitated by a market operator such as the Belgian EPEX SPOT power exchange, which manages electricity trading for Belgium and Luxembourg. Most European markets are further coupled by virtue of participating in Single Day-ahead Coupling (SDAC), which coordinates the clearing of spot markets through the EUPHEMIA (Pan-European Hybrid Electricity Market Integration Algorithm) \cite{euphemia2024} for more efficient allocation of cross-border transmission capacity.

This DAM allows participants to anonymously nominate bids for consumption (or generation) for day $d+1$ before market Gate Closure Time (GCT), typically the noon of day $d$. Participants can offer or demand a certain amount of electricity at a certain price for each hour of the next day, and these supply and demand bids are subsequently matched using a 'pay-as-clear' mechanism to determine a single price per hour for the next day. Shortly after GCT, the market actors get to know the hourly electricity prices and which of their bids were accepted.

The pay-as-clear mechanism sets the clearing price equal to the highest accepted bid \cite{kirschen2018fundamentals}, which, combined with the short time frame of the market and the general inelasticity of electricity demand, can result in large price fluctuations even from one hour to the next on the same day. Increasing variability in supply due to Variable Renewable Energy (VRE) sources \cite{kazmi2022good} as well as accelerating demand electrification (transportation, heating, and industry) \cite{ruhnau2019direct}, further contribute to these swings as well.

These hourly price differences can have significant impacts on consumers in industries where energy consumption makes up a high share of the overall cost structure. As consumers do not know the market clearing price (MCP) at the time when making their demand nominations and might not be flexible in scheduling their consumption, they might regularly pay high prices for the consumed electricity. Downstream effects of elevated and highly variable electricity prices caused large losses in the chemical, paper, and metals industries during the 2021-2023 energy crisis in Europe \cite{European_Central_bank_industrial_production_2023}. Accurate forecasting of prices could prove extremely valuable in these times for consumers able to strategically schedule/shift their electricity consumption to match price patterns, leading to cost savings, derisking of operations, and improved utilization of resources \cite{ayon2017optimal}. Similarly, traders able to accurately forecast price variations can earn large profits by strategically bidding in the market \cite{sunar2019strategic}.

The expansion of cross-border capacity between different bidding zones has increased the participation of actors from neighboring markets. This can lead to a dampening effect on the price fluctuations that occur within and between markets: price convergence or harmonization between different zones and a reduction in price volatility. However, while European markets are increasingly interconnected, individual bidding zones can still exhibit notable differences, due to connection availability constraints etc. These differences also extend to what data is made available for each market, and in the time they are made available, including the GCT and MCP publication. 

\subsection{Electricity Price Forecasting}\label{subsec1.2:Intro_EPF}

The field of Electricity Price Forecasting (EPF) focuses on predicting future electricity prices, and predictions can be made in the long term, such as by a transmission system operator (TSO) to assess the need to attract new generation capacity, but also in the short term, such as by a trader for the spot prices of the next day \cite{Economic_SCheduling_Mathaba_2014}. This latter, short-term EPF, is the focus of this study.

EPF models typically rely on inputs that can help in predicting the prices for the next day, such as previously observed price data and different fundamental market variables that stand in relation to the MCP. Examples for market variables include forecasts for electricity demand \cite{ullrich2012realized} or generation \cite{shi2021effective}, as well as calendar information \cite{zafar2022multistage} and weather data (or forecasts) \cite{sgarlato2022role}. 
Cross-market flows are typically not considered, but features from neighboring markets are sometimes used as covariates to improve the price forecasts in a target market \cite{van2021electricity} (e.g. French net load as a covariate for Belgian DAM price forecasts etc.). It is then evident that using the most accurate input data and forecasts for these covariates is critical for precise predictions. 

In the short-term forecasting of spot prices, researchers have utilized many different types of modeling techniques, ranging from simple time series forecasting methods \cite{shah2021short} to more complex machine learning based techniques. Most statistical techniques depend on a variant of regression analysis, which can additionally be equipped with regularization techniques \cite{friedman_coordinate_descent_lasso_regularization_2010}, allowing many inputs to be used for accurate predictions while managing model dimensionality. Meanwhile, different machine learning methods have also been applied to EPF, including Deep Neural Networks (DNN) \cite{shi2021effective}, random forests and gradient boosted trees \cite{galarneau2023foreseeing}. These methods can learn non-linear relationships between the inputs and predictions for electricity prices, making them highly performant in EPF \cite{j_lago_deep_learning_epf_2018-1}. However, most of these techniques are inherently black-box in nature, i.e. they are not fully interpretable on their own and thus have to be analysed with interpretability techniques to explain their predictions \cite{fan_interpretability_neural_networks_2021}, a pre-requisite to foster trust.

However, there is an inherent trade-off here in the use of complex models with many covariates (typically in the order of hundreds or thousands): prediction models with high computational complexity could prevent price predictions from being generated between when the forecasts for input covariates are made available and the GCT. This could inhibit the strategic placement of bids in the predicted market, and thus indicates a clear trade-off, highlighting the importance of considering time complexity in addition to accuracy and interpretability.

\subsection{Contributions and Paper Structure}\label{subsec1.3:Intro_Paper_Structure}

\begin{figure}[!htb]
    \centering
    \includegraphics[width=0.85\linewidth,]{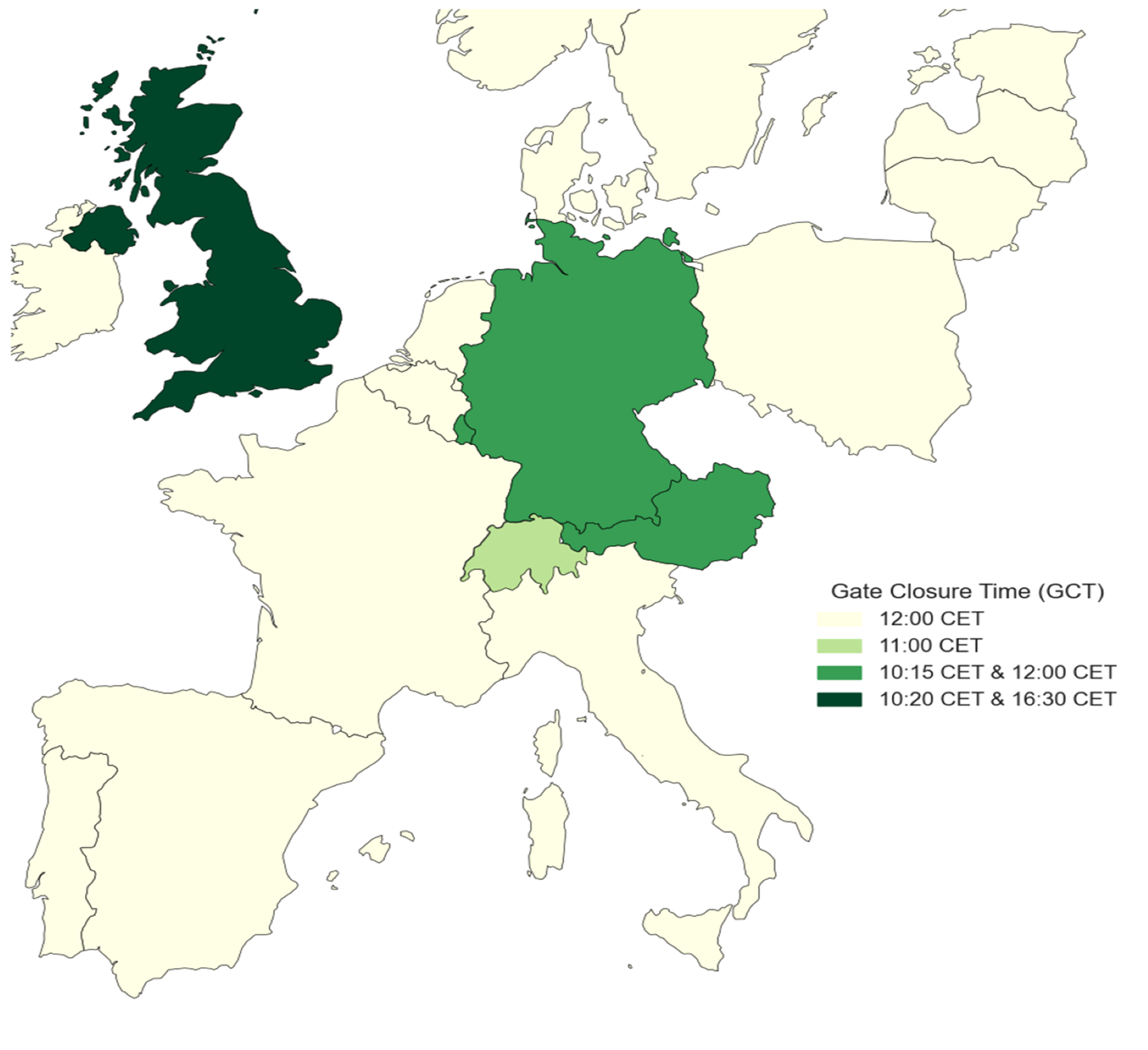}
    \caption{Gate Closure Times across Selected European Markets (Non-Exhaustive Overview: EPEX SPOT, Nord Pool, OMIE, SEMOpx, EXAA).}
    \label{Fig:Map_GCT}
\end{figure}

This study was motivated by a combination of two observations: (1) European spot markets are increasingly coupled in terms of prices, and (2) gate closure times are not unified across bidding zones in Europe. Market price data is often published shortly after GCT. When this happens prior to GCT in another bidding zone, it can be used as an input covariate to forecast prices on this latter bidding zone, theoretically leading to more accurate predictions. Existing differences in GCT between zones in Europe, as the situation stands in early 2025, is shown in Figure \ref{Fig:Map_GCT}.

This motivates our research question: could these differences in the gate closure time between markets be leveraged to improve price forecasting?
This question, which to our knowledge has not yet been explored in previous research, will be addressed by investigating the effect of using published price data from one market as input on the accuracy, time complexity, and interpretability of predictions for a connected market (the three key EPF metrics identified in the previous section). 

We answer this research question using price data from three markets with GCT earlier than noon: the Swiss (CH) market, which is not a part of SDAC and has a GCT at 11:00 CET, and the 15-minute Market Time Units (MTUs) markets for the Austrian (AT) and Germany-Luxembourg (DE-LU) zones, which were introduced to improve market flexibility and renewable integration, with a GCT at 10:15 CET. This price data is used in forecasting the Belgium (BE) and Sweden 3 (SE3) spot market prices. The BE and SE3 markets vary considerably in their dynamics and interconnections to the AT, DE-LU, and CH markets, allowing the effects to be studied in two distinct cases, and leading to better generalization than if only one market were to be considered.

For both markets considered (BE and SE3), this study will investigate forecast accuracy in both normal and extreme operating conditions, by varying several important factors (choice of source market, frequency of model recalibration and extent of historical context etc.). Additionally, interpretability tools will be used to ascertain that models do not simply regurgitate earlier published market prices without considering additional local data, as this could cause great mispredictions of prices during unforeseen or extreme local events. Finally, the study also examines how market data and other model parameters influence the time complexity of the prediction models, which would affect the real-life use case of short-term price predictions (i.e., in case whether the latest published information could not be processed before GCT of the targeted bidding zones). 

This study is thus valuable for market participants and traders, as improved price forecasts can be used to streamline grid operation and generate economic returns, while the insights into the prediction models with and without market data inputs can provide insights for future studies in EPF. For regulators and market operators, the study demonstrates the influences cross-border data flows (in addition to electricity flows) may have on market design.

The rest of this paper is organized as follows: Section \ref{Sec2:Meth} describes the methodology employed in the paper more precisely, as well as the data used in our case studies. Section \ref{Sec3:Results} presents the results, while Section \ref{Sec4:Discussion} discusses the most important insights and limitations of the study. Section \ref{Sec5:Conclusions} concludes the paper.

\section{Methodology and Data}\label{Sec2:Meth} 

In this section, we describe the two forecast models we employ and their different configurations, the evaluation metrics we use to score these models, and the data we use to train and evaluate the forecast models.

\subsection{Forecasting models}\label{subsec2.:Method_Forecast_models}

We employ two state-of-the-art models in this research: (1) a LEAR (LASSO Estimated Auto-Regression (LEAR) model, based on LASSO (Least Absolute Shrinkage and Selection Operator), and (2) a Deep-Neural Network (DNN). Both of the models, adapted from Lago et al. \cite{jlago_epftoolbox_2021_original_paper}, are known to be highly performant for forecasting prices and were already employed for several use cases in literature, such as a study of data augmentation methods \cite{demir_data_augmentation_2021} or explainability techniques \cite{mascarenhas2024bridging}. These models are described in greater detail in this section. The interested reader is referred to existing literature for a more extensive discussion on models for electricity price forecasting \cite{review_of_the_state_of_the_art_in_epf_with_look_future_weron_2014} or forecasting in general \cite{petropoulos_forecasting_2022}.

\subsubsection{The LEAR Model}\label{subsubsec2.:Method_LEAR_model}

The LEAR model is a statistical technique based on regression. Regression models vary wildly in their complexity and predictive ability, and these include, among others, auto-regressive (AR), auto-regressive with exogenous inputs (ARX), auto-regressive integrated moving average (ARIMA), Markov-switching, and long-term seasonal component models. The LASSO, introduced by Tibshirani et al. \cite{tibshirani_regression_1996}, adds a regularization parameter to the Residual Sum of Squares (RSS) function that is minimized to determine the coefficients of this regression model: 

\begin{equation}\label{Eq:LASSO}
min\{\sum_{i=1}^n\left(p_i-\sum_j x_{i j} \beta_{i j}\right)^2+\lambda \sum_{j=1}^p\left|\beta_j\right|\}
\end{equation}

\noindent Where $n$ reflects the data available during the calibration phase, \(p_i\) are the actual prices, \( \beta_{i j}\) are the model coefficients, (\( x_{i j}\)) are the input features or covariates, and $j$ reflects individual features.

Minimizing the prediction error term ensures these coefficients capture the relation between the features and prices, while the regularization introduced in the LASSO penalizes the absolute value of the coefficients. The LASSO operator thus selects only a subset of these input features, setting the other inputs' coefficients to zero. This reduces the model's complexity and computation time, while still ensuring good predictive performance. Such regularized methods provide implicit feature selection, thereby allowing regression models to include more input variables without loss of accuracy or needlessly increasing the model's dimensions and have therefore become popular in recent years since their first use in spot market forecasting \cite{automated_variable_selection_and_shrinkage_uniejewski,ziel_forecasting_lasso_2015}.

The strength of the regularization depends on the selected $\lambda$ value.  Different options exist to determine the optimal lambda value. To determine the optimal lambda value, models might employ a hyperparameter optimization step. This hyperparameter optimization can be based on Cross-Validation (CV) or make use of an information criterion \cite{zou_lars_lasso_2007}, such as the Akaike Information Criterion (AIC) \cite{akaike_aic_1974} or Bayesian Information Criterion (BIC) \cite{schwarz_estimating_bic_criterion_1978}. In this study, we utilize CV because it can offer better performance when the number of features exceeds the number of samples \cite{hastie_elements_2001}. This method operates by dividing the training data into a number of random subsets, used for training or testing the accuracy of the model. These subsets are used to estimate the accuracy of models using the different $\lambda$ values considered. The $\lambda$ with the greatest estimated accuracy is deemed optimal. 

The LEAR model employs the 'Norm' transformation of the data to improve model performance, which scales the data to the interval [0;1]: the values corresponding to 0 and 1 are the minimal and maximal values for that input feature in the training data. Such scaling operations, or transformations of inputs, have been shown to improve model performance \cite{uniejewski_variance_stabilizing_transformations}.

\begin{equation}\label{Eq:Norm_Transform}
\text{Norm}(x) = \frac{x - \min(x)}{\max(x) - \min(x)}
\end{equation}

\subsubsection{The DNN Model}\label{subsubsec2.:Method_DNN_model}

The second model we employ is the DNN model, also presented by Lago et al. \cite{jlago_epftoolbox_2021_original_paper}. The DNN consists of three parts: the input layer, the output layer, and a number of hidden layers in between. Each of these layers includes a certain number of nodes, also called neurons, which map inputs to outputs in conjunction with their weights and activation functions. A trained NN uses these connections to produce output values on each node in the output layer, given input values for the nodes in the input layer. The term "Deep" indicates that the model allows more complex relations between neurons, such as feedback loops, whereas simple NNs, such as multi-layer perceptrons, only allow the forward flow of calculations.

The DNN model also uses a tuning algorithm to determine optimal values for its hyperparameters prior to training. This optimization process is executed using the Tree-structured Parzen Estimator (TPE) algorithm, implemented through the Hyperopt library \cite{HyperOpt_library_bergstra_2013}. TPE tunes the model by searching through a space comprising two primary components: the neural network architecture and training-related hyperparameters, along with the data transform and feature selection mechanism for the inputs. For the DNN architecture, this hyperparameter optimization algorithm decides the number of neurons in each of the two hidden layers, the activation function, the normalization and weight initialization methods, learning rate, and dropout rate (a form of regularization) and regularization strength for feature selection \cite{jlago_epftoolbox_2021_original_paper}. The feature selection works by deciding for each input variable (as specified in Subsection \ref{subsubsec2:Method_input_parameters}) whether to include this variable or not, indicated by a binary value. Two normalization methods are explored for the data transformation: (1) the 'Norm' transform, as detailed in Subsection \ref{subsubsec2.:Method_LEAR_model}, and (2) the 'Norm1' transform, which scales data into the interval [-1, 1].

\subsection{Model Configurations}\label{subsec2:Method_model_configurations}

We test a variety of model configurations for both the LEAR and the DNN model to study the effect of including previously published market data on prediction accuracy, model complexity and interpretability. These model configurations range from choice of input covariates, length of calibration windows, recalibration frequency and model combination via ensembling.

\subsubsection{Choice of input Covariates}\label{subsubsec2:Method_input_parameters}

To evaluate the impact of asynchronous market price data on forecast performance, we used four distinct configurations of input variables for both the Belgian (BE) and Swedish (SE3) electricity markets. These configurations differ slightly for each market, reflecting the specific characteristics of each bidding zone. The input parameters used to predict the BE and SE3 markets under different model configurations, as well as the unit, frequency, and source for this input data, are shown in Table \ref{Tab:Dataset_Electricity_Forecasting}.

\begin{table}[h!]
\caption{Dataset for price forecasting drivers for the Belgian and SE3 electricity spot markets, sourced from Entso-E \cite{entsoe_transparency_platform} and Open-meteo \cite{openMeteo_Site}.}
\label{Tab:Dataset_Electricity_Forecasting}
\centering
\resizebox{\columnwidth}{!}{%
\begin{tabular}{llll}
\hline
 {Variable}                                   &  {Unit}     &  {Frequency} &  {Source} \\ \hline
\multicolumn{4}{l}{{\textbf{Base Configuration BE}}} \\[4pt]
Day-Ahead Load Forecast                           & MW                & Hourly             &  Entso-E \\ 
Wind Forecast (Onshore \& Offshore)& MW                & Hourly             &  Entso-E \\ 
Solar Forecast                              & MW                & Hourly             & Entso-E \\ Temperature Forecast                              & $^{\circ}$C       & Hourly             & Open-meteo \\ 
Humidity Forecast                                 & \%                & Hourly             & Open-meteo \\ 
Day-of-Week Indicator                             & --                & Daily             & National Calendar \\ 
Public Holiday Indicator                         & Binary            & Daily             & National Calendar \\ \hline
\multicolumn{4}{l}{{\textbf{Base Configuration SE3}}} \\[4pt]
Day-Ahead Load Forecast                                     & MW                & Hourly             & Entso-E \\ 
Wind Forecast (Onshore)               & MW                & Hourly             & Entso-E \\ 
Temperature Forecast                        & $^{\circ}$C       & Hourly             & Open-meteo \\ 
Humidity Forecast                        & \%                & Hourly             & Open-meteo \\ 
Day-of-Week Indicator                             & --                & Daily             & National Calendar \\ 
Public Holiday Indicator                          & Binary            & Daily             & National Calendar \\ \hline
\multicolumn{4}{l}{{\textbf{Market Price Data}}} \\[4pt]
DE-LU Spot Price       & €/MWh             & 15-min             & Entso-E \\ 
AT Spot Price                     & €/MWh             & 15-min             & Entso-E \\ 
CH Spot Price                & €/MWh             & Hourly             & Entso-E \\ \hline
\end{tabular}
}
\end{table}

The 'Base' configuration for both zones exclusively includes fundamental market variables and omits any price market data from other countries. The subsequent  'DE/LU, 'DE-LU/AT', and 'DE-LU/AT/CH' configurations sequentially introduce the day-ahead spot market prices from Germany-Luxembourg, Austria, and Switzerland, respectively.

\subsubsection{Choice of calibration Window length}\label{subsubsec2:Method_Cal_windows}

The calibration window length defines the amount of historical data used to calibrate the model and is an important modeling choice. Shorter windows are designed to reflect recent market dynamics, potentially improving responsiveness to recent market changes, while longer windows provide more comprehensive historical contexts for stable coefficient estimation. The calibration windows used for making predictions in this study, across both markets and all four configurations, consist of 56 days (8 weeks), 112 days (16 weeks), 365 days (1 year), and 730 days (2 years). This mix of short and long CWs reflects best practices from published literature \cite{jlago_epftoolbox_2021_original_paper}, and should be an effective mix to capture various degrees of recent trends and long-term market behaviors.  Furthermore, different calibration window lengths provide differing trade-offs between model accuracy and model runtime: longer calibration windows lead to more data processing and thus result in increased training times. In subsequent sections, a model using a certain calibration window is labeled using the identifier 'CWx', where 'x' represents the number of days included in the calibration period.

\subsubsection{Recalibration Frequency}\label{subsubsec2:Method_Recal_Freq}

Next, we investigate how choice of recalibration frequency affects the accuracy and computational efficiency of the forecasting models. As opposed to calibration window length, this affects how often the models are retrained (e.g. daily, weekly etc.) Frequent recalibrations might produce better models (as they are trained on the latest data) but can also lead to much higher computational demands and maintenance requirements. To analyze this trade-off between recalibration frequency and model performance, we evaluate the effect of varying the recalibration frequency for the most accurate ensemble predictions for the BE and SE3 markets. The recalibration intervals explored for these models are daily, weekly, monthly, and, as an extreme scenario, only once at the start of the test period. The evaluation metrics and test period remain the same as other experiments presented in this study. 

\subsubsection{Ensembling}\label{subsubsec2:Method_Ensembling}

This study also utilizes \textit{Ensembling} of several model predictions to obtain a single forecast. This has dual advantages: by combining the predictions, we avoid the impact that one model's specific performance might have, for instance, due to an especially favorable initialization in the DNN model. The reported results will therefore be more representative of the overall effect of the inputs used, such as market data, on the accuracy of the predictions. Second, ensembling is known to result in more accurate overall predictions, predicated on the underlying models making uncorrelated (or even negatively correlated) errors \cite{buizza2019introduction}. The ensemble predictions are generated as the arithmetic average of the forecasts of both models (LEAR and DNN) and four calibration windows (CW56, CW112, CW365, and CW730). Therefore, each ensemble is generated using 8 sets of predictions.

\subsection{Model Evaluation}\label{subsec2:Method_Model_Evaluation}
\subsubsection{Accuracy}\label{subsubsec2:Method_Evaluation_Model_Accuracy}
To evaluate the performance of the predictors, a test period of one year is used for both markets, ranging from 01/01/2024 to 31/12/2024. The training data history depends on the length of the calibration window, but can range as much as two years in the past (walking backwards from the day the prediction is being made for). The accuracy of these predictions is evaluated using five different accuracy metrics:  the Mean Absolute Error (MAE), Root Mean Squared Error (RMSE), relative MAE (rMAE), the symmetric Mean Absolute Percentage Error (sMAPE), and lastly, the R-squared value (R²). The MAE and the RMSE can be calculated as:

\begin{equation}\label{form:MAE}
\mathrm{MAE}
\;=\;
\frac{1}{24\,N_d}
\sum_{d=1}^{N_d}
\sum_{h=1}^{24}
\bigl|p_{d,h}-\hat p_{d,h}\bigr|
\end{equation}

\begin{equation}\label{form:RMSE}
\mathrm{RMSE}
\;=\;
\sqrt{\frac{1}{24\,N_d}
\sum_{d=1}^{N_d}
\sum_{h=1}^{24}
\bigl(p_{d,h}-\hat p_{d,h}\bigr)^2}
\end{equation}

\noindent where \(p_{d,h}\) indicates the actual prices on day $d$ at hour $h$ while \(\hat p_{d,h}\) the predicted prices at the same time. \(N_d\) and $n$ indicate the number of days and number of hours, respectively. The MAE and RMSE are both absolute error metrics, but the RMSE is more sensitive to outliers. Since it is difficult to compare different markets with different prices using these metrics, we also employ two relative error metrics, the rMAE and sMAPE, defined as follow: 

\begin{equation}\label{form:rMAE}
r\mathrm{MAE}
\;=\;
\frac{
\displaystyle\frac{1}{24\,N_d}\sum_{d=1}^{N_d}\sum_{h=1}^{24}\bigl|p_{d,h}-\hat p_{d,h}\bigr|
}{
\displaystyle\frac{1}{24\,N_d}\sum_{d=1}^{N_d}\sum_{h=1}^{24}\bigl|p_{d,h}-\hat p_{d,h}^{\mathrm{naive}}\bigr|
}
\end{equation}

\begin{equation}\label{form:sMAPE}
\mathrm{sMAPE}
\;=\;
\frac{100\%}{24 N_d}
\sum_{d=1}^{N_d}
\sum_{h=1}^{24}
\frac{\bigl|p_{d,h}-\hat p_{d,h}\bigr|}{\bigl(|p_{d,h}|+|\hat p_{d,h}|\bigr)/2}
\end{equation}

In Equation \ref{form:rMAE}, \(\hat p_{d,h}^{\mathrm{naive}}\) denotes the one-week seasonal persistence forecast (the actual price observed exactly seven days earlier). The \(r\mathrm{MAE}\) expresses the model’s MAE relative to this naive benchmark to indicate the added value over simple weekly repetition. The accuracy of these naive predictions is also reported to indicate how difficult the prices are to predict. The second relative metric, the sMAPE, scales the absolute errors with the real price to arrive a percent error. 
Last, the R-squared metric explains what proportion of the variance of the prices can be explained from the variance in the input parameters of the prediction model:

\begin{equation}
R^2 = 1 - \frac{\text{sum squared regression (SSR)}}{\text{total sum of squares (SST)}} 
= 1 - \frac{\sum_d\sum_h\left(p_{d,h}-\hat p_{d,h}\right)^2}{\sum_d\sum_h\left(p_{d,h}-\bar p_{d,h}\right)^2}
\end{equation}\label{form:R_squared}
\\

\noindent\textit{Statistical Significance Tests} 
\vspace{.1cm}

To determine if differences in model accuracy were  actually statistical significant (or random artifacts), we implemented the Giacomini-White [GW] test \cite{giacomini_white_tests_2006}. The null hypothesis of this GW test is that two models have equal conditional predictive accuracy, and rejecting this null implies that one model forecasts significantly better than the other. The results for this test are summarized in “heat map” plots in the form of matrices where rows correspond to benchmark models and columns to competing models. The color of each cell indicates the p-value for rejecting the null hypothesis. Thus, a small p-value (e.g. \(p<0.05\) ) indicated in a certain cell signifies that the null hypothesis can be rejected and the column-model is significantly more accurate than the row-model, whereas higher p-values would suggest the difference in prediction accuracy is not statistically significant. A detailed explanation of the mathematics behind these GW tests is provided in the \ref{appendix_GW_test}. \\

\noindent \textit{Prediction of prices at the extremes} 
\vspace{.1cm}

In addition to reporting these error metrics for the overall test period, these five accuracy metrics are also calculated and reported for the bottom 5th and upper 95th percentiles of historical prices from the one-year test period. These extreme prices are difficult to predict in practice, but simultaneously also contribute most to the gains/savings that can be made by strategically nominating bids.

\subsubsection{Computation Time and Complexity}\label{subsubsec2:Method_Evaluation_Model_Comp_time}

The total computation time required to forecast the full year of predictions is measured for models using different calibration windows and recalibration frequencies. The runtimes and accuracies of these models with different configurations are compared with the corresponding ensemble model runtimes and accuracies to evaluate the trade-off between computational time and predictive performance.
This computational time includes the recalibration, input transformation, and prediction steps for the full year of predictions, divided by the total number of days in the year (366, since 2024 is a leap year). The transformation and prediction steps involve simple computations (multiplications and summations) and therefore do not significantly contribute to the overall runtime. Conversely, the recalibration step constitutes the majority of the computational load. Time measurement of each model started at the beginning of the first training and ended after the final inference. Simulations were run on a server equipped with an Intel Xeon Silver 4210 processor (2.2 GHz, 10 cores).

\subsubsection{Interpretability}\label{subsubsec2:Method_Evaluation_Model_Interpretability}

To explore how using market data influences different models' capability to learn the expected relation between fundamental market variables and predicted prices, we explore the {contribution} of each input parameter in the price predictions. A model that simply copies market data from other zones without considering local parameters risks not being able to account for market-specific events, reducing its reliability. 
The Pearson Correlation Coefficients (PCC) between the predicted and actual prices (BE, SE3), and the market prices for the three markets (CH, AT, DE-LU) are therefore estimated. These correlation values for the market prices can indicate whether accuracy gains from incorporating price data also correspond to higher correlations between these prices and the predicted prices.

We further utilize the 'Absolute Normalized Contribution (ANC)' of each specific input parameter in the final price predictions, since the model utilizes several lags for each covariate (i.e., we are interested in the contribution of entire feature classes such as wind generation, rather than individual features such as wind production at a specific lag). To calculate the ANC, let $N$ be the total number of hourly forecasts ($i=1,\dots,N$) and $\mathcal F$ be the set of all features. For each $i$ and each $f\in\mathcal F$, we define:

\(  x_{i,f} = \text{normalized value of feature }f\text{ at observation }i,\)

\(\beta_{i,f}=\text{regression coefficient for feature }f\text{ at }i \text{as calibrated by the LASSO.}\)

The \emph{instantaneous contribution} of feature $f$ at $i$ then equals 
\[
  c_{i,f} = x_{i,f}\,\beta_{i,f}.
\]

Grouping all hourly instances of each base feature into $M$ groups,$\{G_j\}_{j=1}^M$ , 
(e.g.\ $G_{\text{Price}_{D-1}}=\{\text{Price}_{D-1,H=0},\dots,\text{Price}_{D-1,H=23}\}$). The \emph{grouped contribution} of base feature $j$ at $i$ becomes
\[
  C_{i,j}
  = \sum_{f\in G_j} c_{i,f}
  = \sum_{f\in G_j} x_{i,f}\,\beta_{i,f}.
\]
Finally, the \emph{absolute normalized contribution} of feature $j$ is its average absolute grouped contribution:
\[
  \mathrm{ANC}_j
  = \frac{1}{N} \sum_{i=1}^{N} \bigl|C_{i,j}\bigr|
  = \frac{1}{N} \sum_{i=1}^{N}
    \left|\sum_{f\in G_j} x_{i,f}\,\beta_{i,f}\right|.
\]

This analysis of contributions provides insight into the significance of parameters in the model predictions compared to what is expected for that input parameter based on established domain knowledge. This interpretability analysis further serves as an initial exploration for the deeper investigations in future research that could focus on optimizing both predictive performance and interpretability, particularly in dynamic market environments.

\section{Results}\label{Sec3:Results}

This section presents the results for price forecasting models for the BE and SE3 markets, using the different model configurations discussed in Section \ref{Sec2:Meth}. Subsection \ref{subsec4.1:Forecast_accuracy} examines the forecast accuracy for the BE and SE3 zones, while Subsections \ref{subsec4.2:Results_Computation_Time} and \ref{subsec4.3:Results_Interpretability} discuss the computational time and interpretability of the models.

\subsection{Accuracy}\label{subsec4.1:Forecast_accuracy}

We further split this section into the two bidding zones (BE and SE3) to ensure results can be communicated in a clear manner. Results for each bidding zone follow the same structure: overall accuracy, followed by accuracy during extreme market conditions (lower and upper 5th percentile of observed prices), and finally concluding with statistical significance tests. In each of these categories, results for four model types can be seen: (1) naive prediction (i.e., the weekly persistence model), (2) ensemble base (i.e., the best performing model combination which does not utilize cross-border market information), (3) ensembles with different combinations of cross-border market information (in total four combinations are possible based on the sequence of published prices), and (4) a model that simply uses cross-border prices from a different bidding zone (AT, DE-LU or CH) as its forecast. 

\subsubsection{Belgium (BE)}\label{subsubsec4.1:Forecast_accuracy_BE}

Table \ref{Tab:Ensemble_Forecast_BE} reports the prediction accuracies for the Belgian zone for all prediction models. It shows that all ensemble models demonstrate significant accuracy improvements over the naive prediction benchmark, with the base model improving the MAE from 28.22 to 13.01 €/MWh. Incorporating price data from the Austrian market (ensemble AT) increases accuracy over the base configuration to a MAE of 10.82 €/MWh, while ensemble DE-LU leads to an even lower MAE of 10.22 €/MWh. Combining information from both AT and DE-LU bidding zones (ensemble AT/DE-LU) improves this further to 10.45 €/MWh, with the model containing all cross-border information (ensemble AT/DE-LU/CH) yields the most accurate predictions, achieving a MAE of 10.18. This marks a roughly 22\% reduction in MAE and almost 25\% in RMSE. Copying the market prices from neighboring countries as forecasts (Price CH, Price AT, and Price DE-LU) results in significantly higher accuracy than the naive model but lower accuracy than all the ensemble methods, as expected (including ensemble base). The rMAE score shows the best ensemble model is close to three times more accurate than the baseline naive model.

\begin{table}[ht]
\caption{Prediction accuracies for the ensemble forecasts of the BE bidding zone prices using different sets of market data as input parameters.}\label{Tab:Ensemble_Forecast_BE}

\centering
\small
\resizebox{\columnwidth}{!}{%
\begin{tabular}{lccccc}
\hline
{Forecast Model} & {MAE [€/MWh]} &  {rMAE} &  {RMSE [€/MWh]} &  {sMAPE [\%]} &  {R\textsuperscript{2}} \\
\hline
naive prediction        & 28.22 & 1.00 & 40.99 & 28.79 & 0.09 \\
ensemble base           & 13.01 & 0.46 & 19.68 & 15.83 & 0.79 \\
ensemble AT             & 10.82 & 0.38 & 15.67 & 14.05 & 0.87 \\
ensemble DE-LU          & 10.22 & \textbf{0.36} & \textbf{14.67} & 13.39 & \textbf{0.88} \\
ensemble AT/DE-LU       & 10.45 & 0.37 & 15.13 & 13.58 & \textbf{0.88} \\
ensemble AT/DE-LU/CH    & \textbf{10.18} & \textbf{0.36} & 14.85 & \textbf{13.28} & \textbf{0.88} \\
Price AT                & 16.18 & 0.57 & 23.22 & 17.54 & 0.71 \\
Price DE-LU             & 14.55 & 0.52 & 21.72 & 16.18 & 0.75 \\
Price CH                & 18.12 & 0.64 & 27.26 & 18.55 & 0.60 \\
\hline
\end{tabular}
}
\end{table}  

\vspace{.1cm}
\noindent\textit{Lower 5th\% Prices} 
\vspace{.1cm}

Table \ref{Tab:Ensemble_Forecast_BE_5_percent_prices} shows the accuracies of the ensemble forecast for the lower 5th percentile of prices for the BE zone. Many of the same observations still hold, i.e. naive prediction performs worse than ensemble base, which performs worse than models that incorporate cross-border information. More concretely, compared to naive predictions, using all market data for the predictions (ensemble AT/DE-LU/CH) reduces the MAE from 37.701 to only 12.76 €/MWh, which is only slightly higher than the average accuracy of this model when predicting all prices. Notably, just using DE-LU market prices as the BE forecast achieves the highest accuracy on the lowest 5th\% prices, with an MAE of 12.3 €/MWh, outperforming even the best ensemble model. This is likely explained by the significant effect German renewables can have in pushing down neighboring market prices. This exercise also reveals the greater difficulty in predicting extreme prices: the average forecast errors during these low-price periods are much higher for most model than the average result. \\

\begin{table}[h]
\caption{Prediction accuracies for the ensemble forecasts of the lower 5th percentile of BE bidding zone prices using different sets of market data as input parameters.}\label{Tab:Ensemble_Forecast_BE_5_percent_prices}
\centering
\small
\resizebox{\columnwidth}{!}{%

\begin{tabular}{lccccc}
\hline
{Forecast Model} & {MAE [€/MWh]} & {rMAE} & {RMSE [€/MWh]} & {sMAPE [\%]} & {R\textsuperscript{2}} \\
\hline
naive prediction        & 37.70 & 1.00 & 49.40 & 88.87 & -3.63 \\
ensemble base           & 19.29 & 0.51 & 26.42 & 78.70 & -0.32 \\
ensemble AT             & 15.12 & 0.40 & 21.82 & 71.57 & 0.10 \\
ensemble DE-LU          & 12.68 & 0.34 & \textbf{20.21} & 65.30 & 0.23 \\
ensemble AT/DE-LU       & 13.63 & 0.36 & 21.39 & 67.36 & 0.13 \\
ensemble AT/DE-LU/CH    & 12.76 & 0.34 & 21.22 & \textbf{64.38} & 0.15 \\
Price AT                & 22.32 & 0.59 & 32.63 & 78.77 & -1.02 \\
Price DE-LU             & \textbf{12.30} & \textbf{0.33} & 20.50 & 67.71 & \textbf{0.20} \\
Price CH                & 18.40 & 0.49 & 33.12 & 72.37 & -1.08 \\
\hline        

\end{tabular}
}
\end{table}

\vspace{.1cm}
\noindent\textit{Upper 5th\% Prices} 
\vspace{.1cm}

The forecasting accuracy for the upper 5th percentile of prices is shown in Table \ref{Tab:Ensemble_Forecast_BE_95_percent_prices}. For high-price events, the ensemble forecasts significantly outperform naive predictions once again, as the MAE is reduced from 48.46 to 18.98 €/MWh for the ensemble AT/DE-LU/CH. The notable accuracy gain from incremental addition of market data (from AT, DE-LU, and CH markets) leads to a similar conclusion as for the lower 5th percentile and the full range of prices: the use of price data from additional markets as input parameters increases the forecasting accuracy. However, contrary to the results of the lower 5\%, the best results for the upper 5th\% of prices do not come from using the DE-LU market prices but from the entire ensemble AT/DE-LU/CH. As expected, simply using cross-border prices as market forecasts does not outperform the ensemble models with cross-border information in this case (unlike for the case of low prices).

\begin{table}[h]\caption{Prediction accuracies for the ensemble forecasts of the upper 5th percentile of BE bidding zone prices using different sets of market data as input parameters.}\label{Tab:Ensemble_Forecast_BE_95_percent_prices}
\centering
\small
\resizebox{\columnwidth}{!}{%

\begin{tabular}{lccccc}
\hline
 {Forecast Model} &  {MAE [€/MWh]} &  {rMAE} &  {RMSE [€/MWh]} &  {sMAPE [\%]} &  {R\textsuperscript{2}} \\
\hline
naive prediction        & 48.46 & 1.00 & 72.52 & 16.04 & -0.68 \\
ensemble base           & 30.99 & 0.64 & 52.44 & 8.74  & 0.12 \\
ensemble AT             & 20.66 & 0.43 & 35.17 & 5.88  & 0.61 \\
ensemble DE-LU          & 19.85 & 0.41 & 32.66 & 5.69 & 0.66 \\
ensemble AT/DE-LU       & 20.44 & 0.42 & 34.48 & 5.86  & 0.62 \\
ensemble AT/DE-LU/CH    & \textbf{18.98} & \textbf{0.39} & \textbf{31.66} & \textbf{5.49}  & \textbf{0.68} \\
Price AT                & 25.08 & 0.52 & 37.70 & 6.92  & 0.55 \\
Price DE-LU             & 29.97 & 0.62 & 48.41 & 7.80  & 0.25 \\
Price CH                & 29.52 & 0.61 & 51.15 & 9.65  & 0.17 \\
\hline
\end{tabular}
}
\end{table}

\begin{figure}[!htb]
    \centering
    \includegraphics[width=0.5\linewidth]{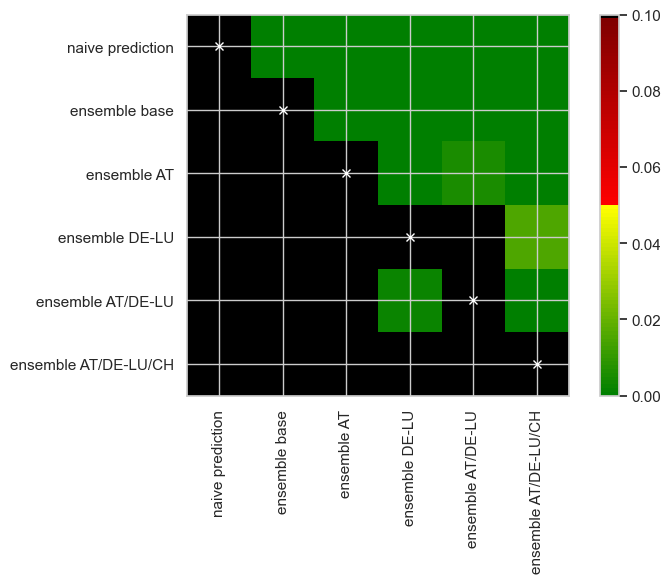}
    \caption{The results of the one-sided Giacomini-White tests for the predictions of the BE prices, made with different sets of market data.}
    \label{Fig:GW_BE}
\end{figure}

Figure \ref{Fig:GW_BE} displays the results for the one-sided GW test for the (overall) predictions for the Belgian bidding zone. In this heat map plot, the color indicates the level of statistical significance for the hypothesis that predictions from the forecast in a column are more accurate than the forecast in a certain row. These results verify that in each subsequent step of adding more market data to the model's input parameters, the ensemble makes predictions for the BE model that are more accurate with significance levels of 5\%.

\subsubsection{Sweden (SE3)}\label{subsubsec4.2:Forecast_accuracy_SE3}

Table \ref{Tab:Forecast_Accuracy_SE3} displays the forecast accuracies for ensemble predictions of the SE3 bidding zone prices for models using different sets of market data as input. Table \ref{Tab:Forecast_Accuracy_SE3} shows that for the SE3 zone, as for the BE zone, the ensemble prediction models using market data significantly outperform naive predictions and the single-region price predictions across all metrics. The best-performing model, ensemble DE-LU, achieves the lowest MAE (10.89 €/MWh), indicating an over 40\% improvement over the naive model and over 9\% compared to ensemble base. This latter is much smaller than what we saw for the BE market.
In fact, in some cases adding more data to the ensemble even leads to a slight decrease in accuracy. This reflects weaker market coupling, as seen in the very poor scores obtained by simply using cross-border market prices as the SE forecast. Nevertheless, a 9\% reduction in MAE and over 13\% in RMSE shows considerable potential for practical utility of the proposed method.

\begin{table}[h]\caption{{Prediction accuracies for the Ensemble Forecasts of the SE3 bidding zone prices using different sets of market data as input parameters.}}\label{Tab:Forecast_Accuracy_SE3}
\centering
\small
\resizebox{\columnwidth}{!}{%

\begin{tabular}{lccccc}
\hline
 {Forecast Model} &  {MAE [€/MWh]} &  {rMAE} &  {RMSE [€/MWh]} &  {sMAPE [\%]} &  {R\textsuperscript{2}} \\
\hline
naive prediction        & 26.41 & 1.00 & 43.60 & 46.77 & -0.19 \\
ensemble base           & 11.79 & 0.45 & 22.66 & 27.24 & 0.68 \\
ensemble AT             & 11.34 & 0.43 & 19.94 & 27.58 & 0.75 \\
ensemble DE-LU          & \textbf{10.89} & \textbf{0.41} & \textbf{19.61} & \textbf{26.85} & \textbf{0.76} \\
ensemble AT/DE-LU       & 11.02 & 0.42 & \textbf{19.61} & 27.04 & \textbf{0.76} \\
ensemble AT/DE-LU/CH    & 11.16 & 0.42 & 20.53 & 27.09 & 0.74 \\
Price AT                & 48.84 & 1.85 & 61.73 & 49.45 & -1.39 \\
Price DE-LU             & 46.43 & 1.76 & 61.59 & 48.16 & -1.38 \\
Price CH                & 44.50 & 1.69 & 56.78 & 47.69 & -1.02 \\
\hline
\end{tabular}
}
\end{table}

\vspace{.1cm}
\noindent\textit{Lower 5\% Prices}
\vspace{.1cm}

Table \ref{Tab:Ensemble_Forecast_SE3_5_percent_prices} shows the forecast performance for the lowest 5\% of SE3 prices. It demonstrates the ensembles’ superior predictive capabilities relative to naive and single-region model predictions for the lower 5th percentile of prices. While adding the AT market price data increases accuracy over the base model, further including data from the DE-LU and CH markets beyond the AT market decreases accuracy (on both MAE and RMSE metrics), suggesting these contain only a limited added benefit for predicting extremely low prices. As observed for the accuracy of predicting the full range of prices for the SE3 market, the direct use of CH, DE-LU, or AT prices shows a drastically lower accuracy than the ensemble forecasting models. Likewise, the very low R2 values indicate that in these extreme conditions, price forecasts actually under-perform even the signal mean.

\begin{table}[h]\caption{Prediction accuracies for the ensemble forecasts of the lower 5th percentile of SE3 bidding zone prices using different sets of market data as input parameters.}\label{Tab:Ensemble_Forecast_SE3_5_percent_prices}
\centering
\small
\resizebox{\columnwidth}{!}{%

\begin{tabular}{lccccc}
\hline
 {Forecast Model} &  {MAE [€/MWh]} &  {rMAE} &  {RMSE [€/MWh]} &  {sMAPE [\%]} &  {R\textsuperscript{2}} \\
\hline
naive prediction        & 18.29 & 1.00 & 25.39 & 89.89 & -11.39 \\
ensemble base           & 6.00  & 0.33 & 9.59  & \textbf{67.94} & -0.77 \\
ensemble AT             & \textbf{5.61}  & \textbf{0.31} & \textbf{8.10 } & 71.36 & \textbf{-0.26} \\
ensemble DE-LU          & 5.76 & 0.32 & 8.27 & 72.03 & -0.31 \\
ensemble AT/DE-LU       & 5.91  & 0.32 & 8.47  & 72.15 & -0.38 \\
ensemble AT/DE-LU/CH    & 6.14  & 0.34 & 8.77  & 72.64 & -0.48 \\
Price AT                & 55.42 & 3.03 & 66.76 & 87.30 & -84.63 \\
Price DE-LU             & 50.63 & 2.77 & 63.43 & 84.65 & -76.29 \\
Price CH                & 55.91 & 3.06 & 70.64 & 84.74 & -94.88 \\
\hline
\end{tabular}
}
\end{table}

\vspace{.1cm}
\noindent\textit{Upper 5\% Prices}
\vspace{.1cm}

Forecast accuracies for the highest 5\% of SE3 prices are presented in Table \ref{Tab:Ensemble_Forecast_SE3_95_percent_prices}, where considerably different behavior compared to lower extremes can be seen. Compared to the base ensemble, model accuracies do increase with the inclusion of AT and DE-LU data but decrease upon adding CH. However, curiously enough, the cross-border price predictors emerge as the strongest models, with especially \textit{Price CH} demonstrating significantly better accuracy than the base models and cross-border ensembles. We leave exploring these results in greater detail from a market fundamentals perspective as a promising future research direction.

\begin{table}[h]\caption{Prediction accuracies for the ensemble forecasts of the upper 5th percentile of SE3 bidding zone prices using different sets of market data as input parameters.}\label{Tab:Ensemble_Forecast_SE3_95_percent_prices}
\centering
\small
\resizebox{\columnwidth}{!}{%

\begin{tabular}{lccccc}
\hline
 {Forecast Model} &  {MAE [€/MWh]} &  {rMAE} &  {RMSE [€/MWh]} &  {sMAPE [\%]} &  {R\textsuperscript{2}} \\
\hline
naive prediction        & 80.14 & 1.00 & 109.67 & 38.59 & -1.45 \\
ensemble base           & 50.26 & 0.63 & 78.99  & 18.50 & -0.27 \\
ensemble AT             & 43.98 & 0.55 & 65.28  & 16.20 & 0.13 \\
ensemble DE-LU          & 42.58 & 0.53 & 65.85 & 15.40 & 0.12 \\
ensemble AT/DE-LU       & 42.25 & 0.53 & 64.80  & 15.15 & 0.14 \\
ensemble AT/DE-LU/CH    & 47.68 & 0.60 & 72.31  & 17.92 & -0.07 \\
Price AT                & 38.27 & 0.48 & 68.22  & 10.41 & 0.05 \\
Price DE-LU             & 36.83 & 0.46 & 73.15  &\textbf{ 9.59}  & -0.09 \\
Price CH                & \textbf{32.19} & \textbf{0.40} & \textbf{61.32}  & 9.74  & \textbf{0.23} \\
\hline
\end{tabular}
}
\end{table}

\begin{figure}[!htb]
    \centering
    \includegraphics[width=0.5\linewidth]{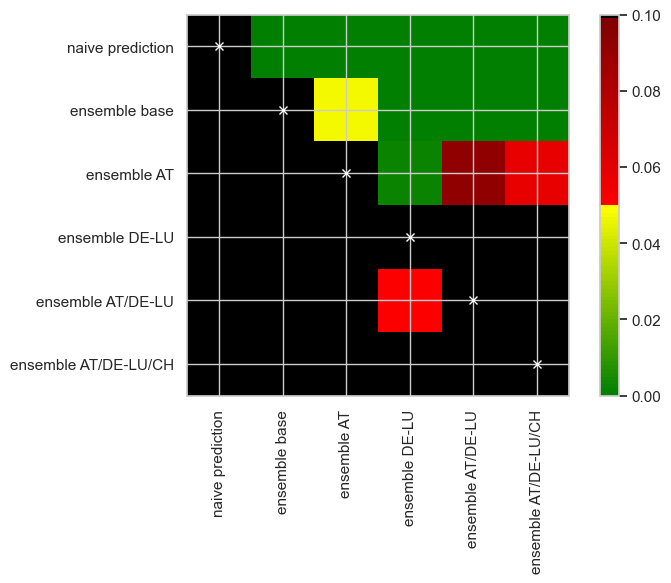}
    \caption{The results of the one-sided Giacomini-White Tests for the predictions of the SE3 prices, using different sets of market data.}
    \label{Fig:GW_SE3}
\end{figure}

The GW statistical tests shown in Figure \ref{Fig:GW_SE3} indicate that models using market price data are significantly more accurate than the base ensemble and naive models for the SE3 zone. However, improvements in accuracy when adding the DE-LU or CH market data to the model using AT market data are not statistically significant below the 5\% significance level, reflecting diminishing returns in forecast improvements. This behavior is very different than what we saw for the Belgian market, and indicates that adding more cross-border data is often but not always better.

\subsubsection{Temporal Evolution of Forecast Accuracy}\label{subsubsec:Comparison SE3vsBE}

\begin{figure}[!htb]
    \centering
    \includegraphics[width=0.6\linewidth]{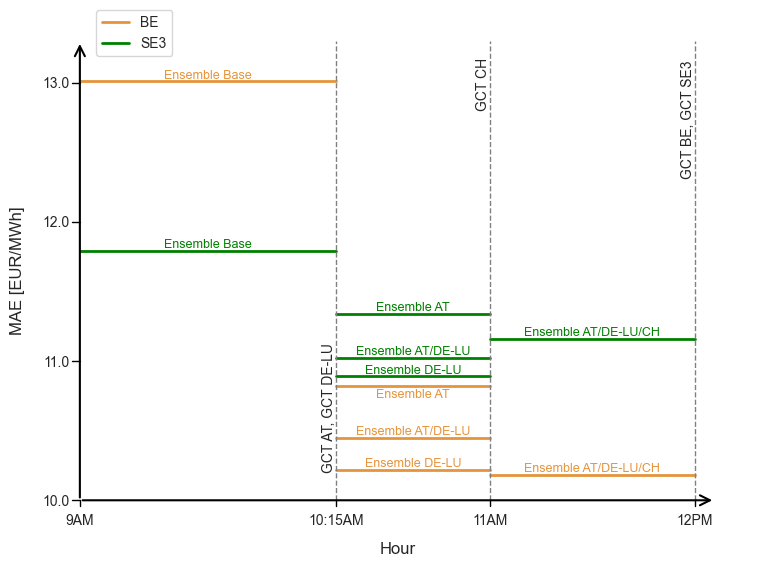}
    \caption{Evolution of forecasting accuracy (MAE) for BE and SE3 DA market prices as additional market price information becomes available at respective GCT.}
    \label{Fig:MAEvsGCT}
\end{figure}

Figure \ref{Fig:MAEvsGCT} illustrates how forecasting accuracy (MAE) can improve as additional day-ahead market price information becomes available, summarizing the results of the previous sections. Without any DA price information from other markets, fundamental variables data can be used to generate the ensemble base predictions,  resulting in MAE values of 13.01 EUR/MWh for BE and 11.79 EUR/MWh for SE3. At 10:15 AM, following the GCT and subsequent publishing of prices from the 15-minute DA markets in the Austrian and Germany-Luxembourg zones, ensemble models incorporating these prices (ensemble AT and ensemble AT/DE-LU) can significantly reduce prediction errors. For Belgium, a further (small) improvement in accuracy is observed after incorporating DA prices from Switzerland, which has a market GCT at 11:00 AM. In the case of SE3, the inclusion of Swiss prices only improves accuracy compared to the ensemble AT, but does not show improvement over the other two models. The ensemble DE-LU model achieves the lowest MAE for SE3. 

The lowest forecast error for SE3 is therefore achieved at 10:15 AM, and for Belgium at 11:00 AM, before the closure of these markets at noon. These findings therefore underline the importance of computational complexity in forecasting: while forecasts become more accurate with additional market information closer to GCT, the practical utility of late forecasts may decrease (owing to the need to schedule demand or generation based on forecast prices), highlighting the trade-off between accuracy and operational feasibility.

\subsection{Computation Time}\label{subsec4.2:Results_Computation_Time}

\begin{figure}[htbp]
    \centering
    \begin{subfigure}[b]{0.48\textwidth}
        \centering
        \includegraphics[width=\textwidth]{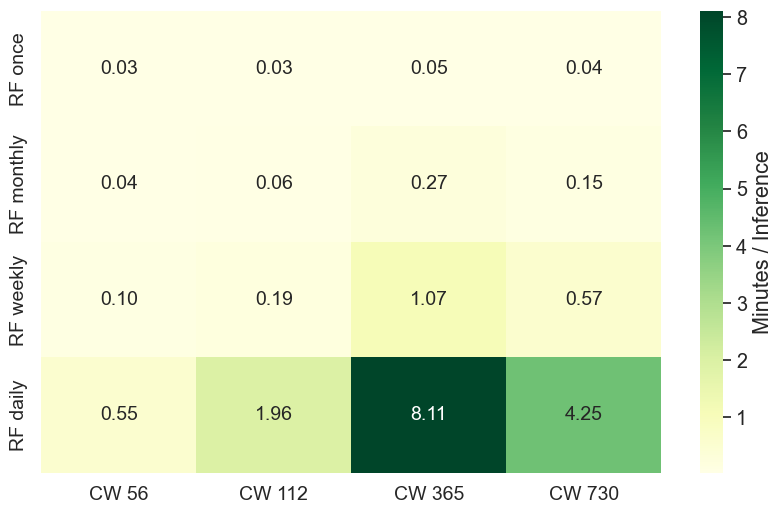} 
        \caption{}
        \label{fig:image_a}
    \end{subfigure}
    \hfill
    \begin{subfigure}[b]{0.48\textwidth}
        \centering
        \includegraphics[width=\textwidth]{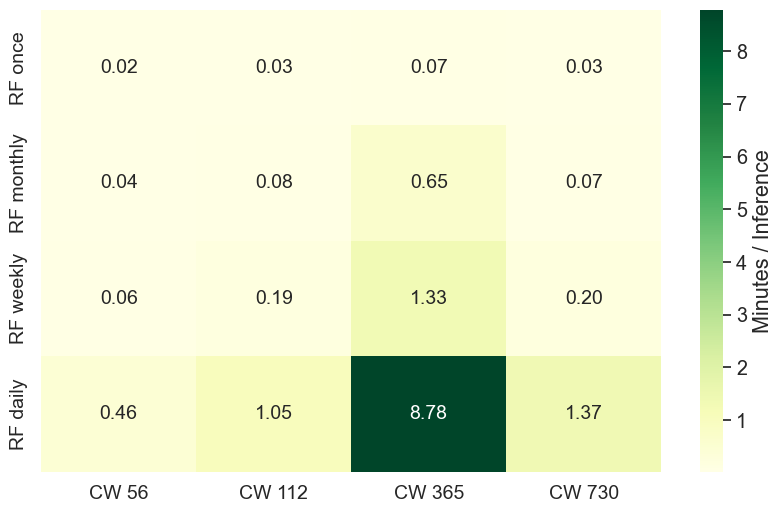} 
        \caption{}
        \label{fig:image_b}
    \end{subfigure}
    
    \caption{The average runtime (minutes) to train and generate predictions across different CWs and RFs with the (a) \textit{LEAR AT/DE-LU/CH} and (b) \textit{LEAR DE-LU} ensembles models for the Belgian and SE3 zones, respectively}
    \label{fig:comp_time}
\end{figure}

The temporal evolution of forecast accuracy with more data availability provides a natural segue to questions about computation time of the ensemble of models. The effect that model parameters have on the computational complexity for different models is explored for the LEAR model of the most accurate ensembles for the BE zone \textit{(LEAR AT/DE-LU/CH)} and the SE3 zone \textit{(LEAR DE-LU)}. Recalibrating daily leads to the most accurate predictions and was used for previous results, but this may require significantly more computational cost and a higher runtime compared with less frequent recalibrations (e.g., weekly, monthly, or yearly). Figure \ref{fig:comp_time} shows the computation time for the LEAR models from the most accurate ensembles, identified by the calibration window (CW), for different recalibration frequencies (RF). Tables \ref{Tab:Recalibration_Frequency_best_model_BE} and \ref{Tab:Recalibration_Frequency_best_model_SE3} report the corresponding accuracy metrics for the ensemble predictions made from these LEAR models. Note that generating these ensemble predictions for each of the RFs would be constrained by the single model with the greatest computation time in that ensemble, assuming running different models is parallelized.

Figure \ref{fig:comp_time} shows the expected pattern: as both recalibration frequency and calibration window size increase, computational time likewise increases. This follows from the higher amount of training data per calibration (higher CW) or the more frequent training (higher RF), leading to increased computational runtime. However, CW 365 shows an exception, as the runtimes for this window are unexpectedly higher than at CW 730, an anomaly that could stem from implementation details or data characteristics (e.g. data for several seasonal periods, as in CW 730, could help model convergence). Importantly, even at the daily RF, the runtime to generate LEAR ensemble predictions remains in the order of seconds to minutes. While this is still within practical limits for the one-hour window between the publishing of market data and 12 pm GCT, this stresses the importance of considering time complexity, as using equipment with lower computation power may risk the computation time (both to generate the forecast and the market bid) exceeding the one-hour limit. 

Tables \ref{Tab:Recalibration_Frequency_best_model_BE} and \ref{Tab:Recalibration_Frequency_best_model_SE3} show that, as expected, lowering the recalibration frequency leads to a decrease in accuracy. However, a decrease in accuracy of less than 8\% when reducing RF from daily to weekly for the BE zone could reduce the calibration time by nearly an order of magnitude for time critical applications. For the SE3 zone, a change from daily to weekly recalibration similarly only results in a small increase in the MAE of 0.17 €/MWh but could again lower the total computation time by nearly sevenfold. Only at very low RFs do the errors increase significantly, but this effect shows differences between the two zones. In the BE zone, the yearly (once) recalibration underperforms even a simple naive model, while for the SE3 zone it still yields acceptable results.

The results in Tables \ref{Tab:Recalibration_Frequency_best_model_BE} and \ref{Tab:Recalibration_Frequency_best_model_SE3} are not for the overall ensemble as in the previous section, but only for the LEAR models, as these are also the models that will be analyzed for interpretability. We observe that, perhaps counter-intuitively, the LEAR ensemble showed slightly lower MAE (±2.5\%) than the corresponding ensemble for the BE bidding zone containing both the LEAR and DNN models displayed in previous results. However, the overall ensemble demonstrated very similar RMSE and R2 scores, leading us to conclude that these differences are not too important.

\begin{table}[h!]\caption{The effect of the recalibration frequency on the prediction errors for the \textit{LEAR AT\_DE-LU\_CH} ensemble model for the BE zone}\label{Tab:Recalibration_Frequency_best_model_BE}
\centering
\small
    \resizebox{\columnwidth}{!}{%

\begin{tabular}{lccccc}
\hline
 {Forecast Model} &  {MAE [€/MWh]} &  {rMAE} &  {RMSE [€/MWh]} &  {sMAPE [\%]} &  {R\textsuperscript{2}} \\
\hline
naive prediction                        & 28.22 & 1.00 & 40.99 & 28.79 & 0.09 \\
RF once         & 44.44 & 1.58 & 73.89 & 35.66 & -1.95 \\
RF monthly     & 12.17 & 0.43 & 20.96 & 14.40 & 0.76 \\
RF weekly       & 10.58 & 0.38 & 16.12 & 13.73 & 0.86 \\
RF daily         & \textbf{9.92}  & \textbf{0.35} & \textbf{14.88} & \textbf{13.08} & \textbf{0.88} \\
\hline
\end{tabular}
}

\end{table}

\begin{table}[h!]\caption{The effect of the recalibration frequency on the prediction errors, for the \textit{LEAR DE-LU} ensemble for the SE3 zone}\label{Tab:Recalibration_Frequency_best_model_SE3}
\centering
\small
    \resizebox{\columnwidth}{!}{%

\begin{tabular}{lccccc}
\hline
 {Forecast Model} &  {MAE [€/MWh]} &  {rMAE} &  {RMSE [€/MWh]} &  {sMAPE [\%]} &  {R\textsuperscript{2}} \\
\hline
naive prediction          & 26.41 & 1.00 & 43.60 & 46.77 & -0.19 \\
RF once  & 15.02 & 0.57 & 22.19 & 34.99 & 0.69 \\
RF monthly & 12.53 & 0.47 & 21.34 & 29.36 & 0.71 \\
 RF weekly  & 11.72 & \textbf{0.44} & 20.78 & 28.13 & \textbf{0.73} \\
RF daily    & \textbf{11.55} & \textbf{0.44} & \textbf{20.65} & \textbf{27.69} & \textbf{0.73}\\
\hline
\end{tabular}}

\end{table}

\FloatBarrier

\subsection{Interpretability}\label{subsec4.3:Results_Interpretability}

In this section, we explore model predictions through two interconnected lens to see if models have learnt anything beyond mimicking already published prices. In the first, we look at the Pearson correlation coefficient between the different markets and produced forecasts. In the second, we take a deeper dive into model feature importances.

\begin{figure}[htbp]
    \centering
    \begin{subfigure}[b]{0.48\textwidth}
        \centering
        \includegraphics[width=\textwidth]{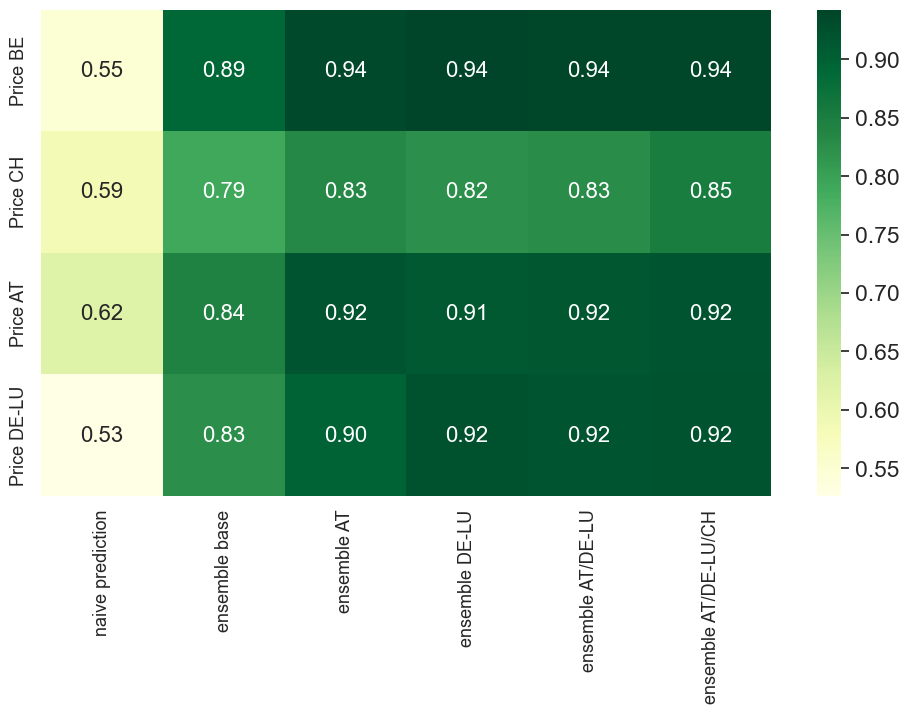} 
        \caption{}
        \label{fig:Correlations_prices_vs_model_predictions_BE}
    \end{subfigure}
    \hfill
    \begin{subfigure}[b]{0.48\textwidth}
        \centering
        \includegraphics[width=\textwidth]{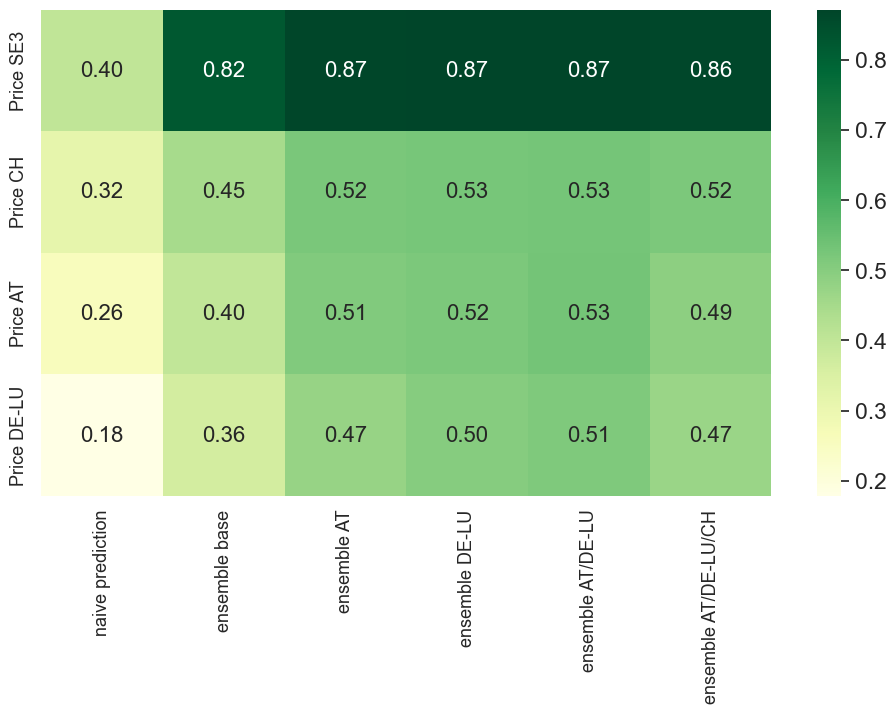} 
        \caption{}
        \label{fig:Correlations_prices_vs_model_predictions_SE3}
    \end{subfigure}
    
    \caption{The Pearson correlation coefficients for the price predictions and the market prices from the AT, DE-LU, and CH zones, with the Belgium (a) and SE3 (b) spot prices.}
    \label{fig:correlation_test}
\end{figure}

Figure \ref{fig:correlation_test} displays the Pearson correlation test values for the ensemble price predictions for the Belgium and SE3 zones with the actual prices of that zone and the AT, DE-LU, and CH zones. The correlation values parallel the accuracies of the ensemble models. Overall, the higher accuracies for the Belgium zone correspond to higher correlation values, and the greater correlation between the naive prediction for the BE prices than for the SE3 prices might further suggest that the BE zone is more easily predictable.
The much greater correlations for the ensemble predictions using market data and the actual AT, DE-LU, and CH prices for the Belgium zone than for the SE3 zone indicate that the models for Belgium show greater reliance on market data and lesser reliance on other input parameters, compared to the models for SE3. 

\begin{figure}[!htb]
    \centering
    \includegraphics[width=0.5\linewidth]{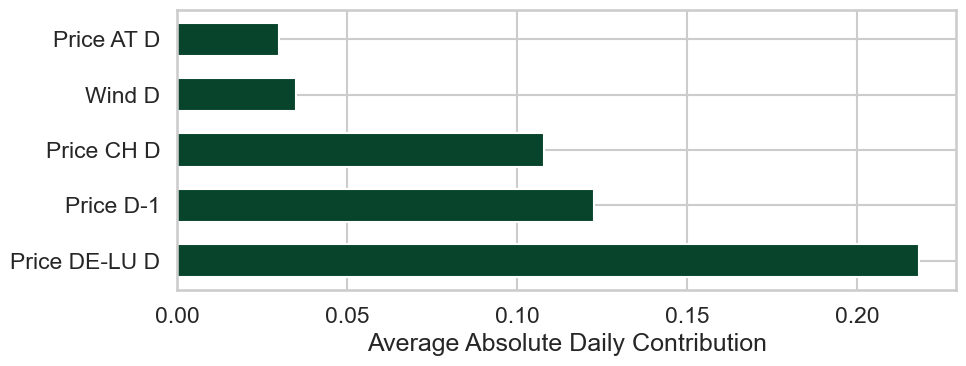}
    \caption{Average ANCs for each of the five most significant features for the BE zone.}
    \label{Fig:Averaged_absolute_contributions_BE}
\end{figure}

\begin{figure}[!htb]
    \centering
    \includegraphics[width=0.5\linewidth]{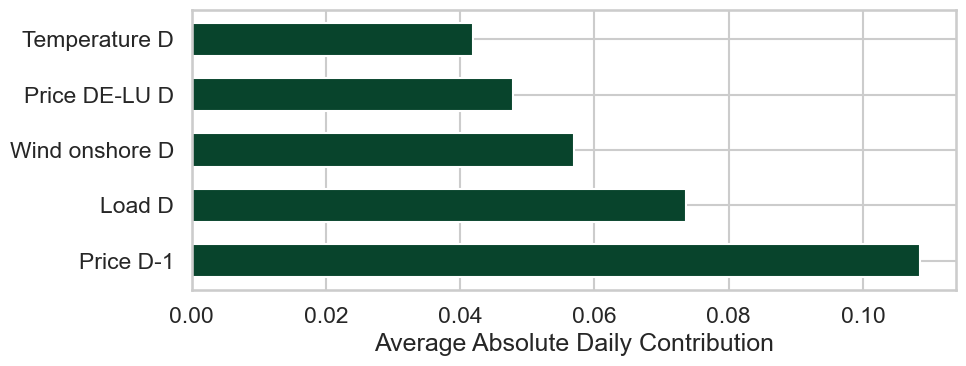}
    \caption{Average ANCs for each of the five most significant features for the SE zone.}
    \label{Fig:Averaged_absolute_contributions_SE3}
\end{figure}

To better understand the importance given to different input features, we take a closer look at the most accurate single {LEAR} models for the Belgium (LEAR CW365 AT/DE-LU/CH) and SE3 (LEAR CW365 AT/DE-LU/CH) zones. The average ANCs from the five input features with the greatest contributions for the full-year price predictions are displayed in Figures \ref{Fig:Averaged_absolute_contributions_BE} and \ref{Fig:Averaged_absolute_contributions_SE3}, ranked from the least significant to the most significant of the five parameters. Subscripts indicate the time of the feature as compared to the day of the predicted prices (\(d\)).

Results for the top features align with expectations based on market dynamics and interconnections. The \(Price\ DE-LU\ D\) is the most influential feature for the BE zone, while only the fourth most important for the SE3 market, reflecting the closer tie of the German-Luxembourg zone with Belgium’s electricity system due to greater cross-border trading and market coupling. 
Temporal dependencies (\(Price\ D-1\)) are significant for both zones, as DA electricity prices generally exhibit high autocorrelation, making past prices a strong predictor of future ones.
The features \(Price\ CH\ D\) and \(Price\ AT\ D\) indicate that, even with the inclusion of the DE-LU zone prices, the AT and CH prices still add to the overall predictive value of the model for the BE zone, which is, however, not reflected in the SE3 zone. 
The \(Wind\ D\) represents the only fundamental parameter not corresponding to a type of price data for Belgium, while three fundamental parameters, \(Wind\ onshore\ D\), \(Load\ D\) and \(Termperature\ D\), are included in the top features for the SE3 zone.

These results indicate that the model for BE attributes much greater importance to the DA electricity prices of other countries than the SE3 zone, suggesting that prediction accuracy improves more from the inclusion of market data from a zone to which it is more closely tied. These results thus reflect the continuing integration of European electricity markets, where market coupling and cross-border trade have made price formation increasingly interconnected. 
While leveraging these differences in gate-closer time between bidding zones that show great interconnectivity can help achieve state-of-the-art forecasting accuracies, this can simultaneously be a double-edged sword: heavy reliance on price data, such as in the BE predictions, might also inhibit a model from learning the fundamental market dynamics between supply and demand, which determine the resulting spot prices, creating the risk of great mispredictions of prices during unforeseen or extreme local
events. Nevertheless, our analysis did not unearth any evidence of this during the test period of 2024.

\section{Discussion and Future Work}\label{Sec4:Discussion} 

In this section, we provide a brief overview of the most important findings, and conclude with some limitations and future research directions.

\subsection{Asynchronous Cross-Border Market Price Data Improves Overall Forecast Accuracy}

The results demonstrate that integrating price data from (neighboring) markets can significantly improve electricity price forecasting accuracy. This aligns with the hypothesis that market coupling and increasing interconnectivity between European electricity markets allow published day-ahead prices in one market to enhance forecasts in another.

The magnitude of accuracy improvement depends on the geographical and operational proximity of the input market to the forecasted market. For the Belgian (BE) market, incorporating data from the German-Luxembourg (DE-LU) market, which has strong interconnection and operational similarities, showed great improvements. Subsequently adding data from Austria (AT) or Switzerland (CH) to the input parameters further increased accuracy with each additional market used for the BE zone, as statistically validated by Giacomini-White tests, but the marginal benefit decreased notably with each additional market. Similar observations were made for the SE3 market, where incremental accuracy improvements greatly diminished when adding markets with less direct connectivity, sometimes even leading to small reductions in accuracy. These findings indicate that adding market data from less relevant zones does not always contribute meaningfully to increased predictive accuracy beyond a certain point.

\subsection{Impact on Extreme Price Predictions paints a positive but more complex picture}

Our analysis revealed that the integration of market price data, when carefully selected, can also offer pronounced advantages, specifically during periods of extreme prices. This is a crucial finding given that accurately forecasting extreme price events (both low and high extremes) carries great economic implications for market participants, such as industries with substantial electricity consumption or traders seeking arbitrage opportunities.

For the Belgian market, directly using the published DE-LU market prices as the forecast showed great accuracy in predictions for forecasting the bottom 5\% prices, even slightly outperforming the best ensemble model. However, an ensemble forecast outperformed this for the top 5\% prices, underscoring the advantage of using multiple data sources to capture complex dynamics during extreme high- and low-price periods.
For the SE3 market, however, adding Swiss market data negatively impacted accuracy for extreme price scenarios, reinforcing the idea that not all additional market data necessarily improves predictive capability. This indicates that the efficacy of market data inputs depends heavily on specific market characteristics, such as generation mix, interconnection capacities, and regulatory frameworks.

\subsection{Calibration Window and Recalibration Frequency Offer Important Trade-offs}

The study of the model's computational time in relation to calibration window (CW) length and recalibration frequency (RF) provides additional important insights. The expected pattern was generally followed: higher CWs and RFs led to greater average computation times. Thus, while daily recalibrations produced the most accurate forecasts, this also imposed significantly greater computational burdens. For example, transitioning from weekly to daily recalibrations in the SE3 market improved prediction accuracy only marginally while substantially increasing computational demands, highlighting the importance of considering this trade-off in practice. 

At the same time, incorporating cross-border market data pushes DAM price forecasts closer to GCT in their own respective markets, which can be a challenge if additional complex downstream optimization processes need to be run afterwards to schedule demand or generation. Our results show that while computation times can vary by several orders of magnitude, they still remain manageable in general.

\subsection{(Over-)Reliance on Market Data can Lead to Potential Risks}

Results of the interpretability analysis illustrate that forecasting models can significantly rely on market price inputs from interconnected markets, such as prices from DE-LU for the Belgian zone. For this zone, three of the five most influential input variables corresponded to day-ahead prices from neighboring bidding zones (in addition to the wind forecast and lagged prices in Belgium), potentially overshadowing other locally relevant inputs like solar or demand forecasts.

While this reliance logically follows from market coupling, it also introduces the risk that models might produce overly confident forecasts by "copy-pasting" previously published market data in other bidding zones, especially when extreme and unforeseen market divergence occurs. Such divergence could happen due to localized events (e.g., infrastructure failures, regulatory changes) that may not be captured if models undervalue local variables. This shows a trade-off: exploiting asynchronous market information enhances accuracy under typical market conditions as well as the upper and lower 5th percentile of prices in 2024, but might still lead to forecasting blind spots or overconfidence during truly atypical market events.

\subsection{Limitations and Future Outlook}

Despite the promising results and insights provided, this study is not without its limitations. Two state-of-the-art models were utilized to generalize findings, but future work might consider additional forecasting methods. Likewise, this study focused on the BE and SE3 bidding zones, using markets with earlier gate closures as input sources, but extending the analysis to predict prices in other European markets would further validate and generalize our conclusions. Likewise, this study focused on the year 2024, and thus extending it to other years could further validate the findings. Nonetheless, to the best of our knowledge, this study is the first to explore the use of cross-border asynchronously published market data for improved price forecasting, while giving consideration to the applicability of the use case and simultaneously considering model interpretability and versatility.  

\section{Conclusions}\label{Sec5:Conclusions} 

To conclude, this study examined how differences in the gate closure times between bidding zones could be leveraged to improve electricity price forecasting. This was motivated by the European spot markets becoming increasingly coupled in recent years, such as through the expansion of transmission capacity and automatic transferring of bids, and leading to greater price convergence, while simultaneously still displaying differences in spot market operation, such as in the GCT. As price data might hold information not yet captured by regular market drivers, prices published for a neighboring zone with earlier GCT might add great predictive value for the forecast of a second zone (with later GCT). However, the use of this price data also restricts the time complexity of the prediction models, as the computation time should not exceed the differences between the publishing of prices from the first zone and the GCT of the second zone. 

This was explored in a use case involving both Belgian and SE3 spot prices in 2024, which were predicted while gradually adding price data from the Austria, Germany–Luxembourg, and Switzerland bidding zones to a base set of input covariates. This additional price information led to a maximum improvement in the MAE of 2.83 EUR/MWh or 21.8\% for the Belgian zone and 0.9 EUR/MWh, or 9.2\% for the SE3 zone. In terms of RMSE, even larger improvements were seen (24.5\% and 13.5\% for BE and SE3 respectively). The accuracy gains for the extreme prices in the Belgian zone were even more drastic, as the MAE for the upper 5th percentile and lower 5th percentile could be decreased by 38,8\% and 33.9\% , respectively. As before, these gains were also visible in Sweden, but to a smaller extent. The examination of the coefficients learned by the prediction models indicated that the forecasting models (especially for the BE zone) come to rely greatly on the price data for the predictions (while also still considering other factors such as historical prices and renewable generation forecasts).

In doing so, the forecasting models presented in this paper exhibit truly remarkable gains in accuracy, which can make a huge difference in terms of profitability while trading on the electricity markets.
As such, this study demonstrates that market data can be of great value in improving price forecasting, but modelers should ensure that it is utilized in combination with other fundamental market drivers to generate price forecasts and price data is only leveraged in such cases where it can significantly contribute, such as for predicting prices of zones that are closely related.

\appendix\section{Giacomini-White tests}
\label{appendix_GW_test}


This appendix details the calculation of the GW statistical significance tests. This test indicates the confidence with which a certain prediction can be said to be more accurate than a second prediction. It utilizes a loss-differential series, defined as the difference in the error series made by a first model and that of a second model. This study used the absolute value of the loss series for the GW test, while a combined comparison was performed for the predictive abilities for all 24 hours of the day at once:

\begin{equation}
d_{d, h}^{\mathrm{A}, \mathrm{B}}=L\left(|\varepsilon_{d, h}|^{\mathrm{A}}\right)-L\left(|\varepsilon_{d, h}|^{\mathrm{B}}\right)
\end{equation}

The GW test, shown in Formula \ref{GW_test}, includes the lagged loss differential values $\mathbb{X}_{d-1}$ and tests the conditional predictive ability. The null hypothesis for the one-sided GW test states that $\phi^{\prime} <= 0$. This signifies that when knowing the previous values of the loss differential series, the error of model A is expected to be smaller than that of the second test.

\vspace{3mm}
\begin{equation}
d_d^{\mathrm{A}, \mathrm{B}}=\phi^{\prime} \mathbb{X}_{d-1}+\varepsilon_d
\label{GW_test}
\end{equation}

For this hypothesis, the p-value is calculated. If the p-value is below a certain statistical significance level, usually taken at 5\%, the hypothesis is rejected. This indicates that the first model's errors are larger than those of the second model. Alternatively, a p-value higher than the significance level does not allow the hypothesis to be rejected.




 \bibliographystyle{elsarticle-num} 
 \bibliography{cas-refs}

\end{document}